\documentclass{IOS-Book-Article}
\usepackage{mathptmx}
\usepackage{lineno,hyperref}
\usepackage{xspace}
\usepackage{enumerate}
\usepackage{amssymb}
\usepackage{amsmath}
\usepackage{graphicx}
\usepackage{relsize}
\usepackage{color,colortbl}
\usepackage{csquotes}
\usepackage{amssymb}
\usepackage{graphicx}
\usepackage{xspace}
\usepackage{dsfont}
\usepackage{textcomp}
\usepackage{url}
\usepackage{array}
\usepackage{pdflscape}
\usepackage{rotating}
\usepackage{blindtext}
\usepackage{longtable}
\usepackage{algpseudocode}
\usepackage{caption}
\usepackage{appendix}
\usepackage{afterpage}

\newcommand{\AForm}{\mbox{\ensuremath{\mathsf{A} = \langle \Slabel,\le, \OR,\OA, \OC,\IES, \IE \rangle}}}

\newcommand{\Labelplus}[2]{\mbox{\ensuremath{\mu_{#1}^{#2}}}}
\newcommand{\Labelminus}[2]{\mbox{\ensuremath{\delta_{#1}^{#2}}}}

\newcommand{\nega}[1]{{\ensuremath{\sim}\xspace}#1}
\newcommand{\negaBar}[1]{\ensuremath{\overline{#1}}}

\newcommand{\Sup}{\ensuremath{\odot}}
\newcommand{\Con}{\ensuremath{\ominus}}
\newcommand{\Agr}{\ensuremath{\oplus}}

\newcommand{\neutro}{\ensuremath{\bot}}
\newcommand{\neutros}{\ensuremath{\top}}

\newcommand{\ALL}{\ensuremath{\mathcal A}}
\newcommand{\Slabel}{\ensuremath{A}}
\newcommand{\R}{\ensuremath{\mathcal R}}
\newcommand{\K}{\ensuremath{\mathcal K}}
\newcommand{\F}{\ensuremath{\mathcal F}}

\newcommand{\OR}{\ensuremath{\odot}}
\newcommand{\E}{A}
\newcommand{\OA}{\ensuremath{\oplus}}
\newcommand{\OC}{\ensuremath{\ominus}}
\newcommand{\IE}{\ensuremath{\bot}}
\newcommand{\IES}{\ensuremath{\top}}

\newcommand{\Graph}[1]{\ensuremath{G}}

\newcommand{\LGraph}[1]{\ensuremath{G_{\mathcal{A}}}}
\newcommand{\CLGraph}[1]{\ensuremath{G_{R}}}

\newcommand{\LAFARG}{\mbox{\ensuremath{\langle {\mathcal L}, {\mathcal R},  {\mathcal K}, {\mathcal A}, \F_{\K} \rangle}}}

\newcommand{\LLa}{\ensuremath{{\mathcal L}}}

\newcommand{\facto}[1]{\ensuremath{#1}} 
\newcommand{\drule}[2]{\mbox{\ensuremath{#1 \at #2}}}
\newcommand{\at}{\ensuremath{\,-\!\!\!\prec\,}}
\newcommand{\defleftarrow}{{\raise1.5pt\hbox{\tiny\defleft}}}
\newcommand{\defleft}{\mbox{---\hspace{-1.5pt}\raise.05pt\hbox{$<$}}}
\newcommand{\argdrule}[2]{\drule{#1}{#2}}
\newtheorem{Definition}{{\bf  Definition}}
\newtheorem{Example}{{\bf  Example}}

\def\hb{\hbox to 10.7 cm{}}

\begin{document}

\pagestyle{headings}
\def\thepage{}

\begin{frontmatter}             

\title{Dealing with Qualitative and Quantitative Features in Legal Domains.}

\markboth{}{JURIX 2018\hb}

\author[A,B]{\fnms{Maximiliano C. D. Bud\'an} \snm{}%
},
\author[D]{\fnms{Mar\'ia Laura Cobo} \snm{}},
\author[A]{\fnms{Diego I. Mart\'inez} \snm{}},
and
\author[C]{\fnms{Antonino Rotolo} \snm{}}

\runningauthor{Bud\'an et al.}
\address[A]{Inst. for Computer Science and Eng. (UNS-CONICET),
            Dept. of Computer Science and Engineering,
            Universidad Nacional del Sur, Argentina.}
\address[B]{Dept. of Mathematics, 
            Universidad Nacional de Santiago del Estero, Argentina.}
\address[C]{Dept. of Mathematics, 
	University of Bologna, Italy.}
\address[D]{Dept. of Computer Science and Eng.,
	Universidad Nacional del Sur, Argentina.}

\begin{abstract}
In this work we enrich a formalism for argumentation by including a formal characterization of features related to the knowledge, in order to capture proper reasoning in legal domains. 
We add meta-data information to the arguments in the form of labels representing quantitative and qualitative data about them.
These labels are propagated through an argumentative graph according to the relations of support, conflict, and aggregation between arguments. 
\end{abstract}

\begin{keyword}
Commonsense Reasoning \sep Argumentation Process \sep Qualitative and Quantitative Feature \sep Legal Domain
\end{keyword}

\end{frontmatter}
\vspace{-20pt}

\markboth{JURIX 2018 -- DOI 10.3233/978-1-61499-935-5-176\hb}{JURIX 2018 -- DOI 10.3233/978-1-61499-935-5-176\hb}

\section{Introduction}\label{Sec.Intro}



Argumentation theory specializes in modeling the process of human reasoning to determine which conclusions are acceptable in a context of disagreement ~\cite{BesnardHunter2008,grs:book}.  
The basic idea is to identify arguments in favor and against a claim, then select the acceptable ones through a proper analysis, determining whether the original statement can be accepted or not.
In general, arguments are self-contained logical constructions of defeasible deduction, and dialectical procedures of justification are defined on these constructions. 
There are, however, some important knowledge related to the use, nature and role of arguments that must be considered when analysing a particular case.   
In legal domains, there exist different non-logical features that represent different uncertain dimension .
The aim of this work is to propose a legal-inspired enrichment of argumentation frameworks by using decorations to arguments in the form of \textit{labels}.
These labels are integrated and transformed through the argumentation analysis according to its intended meaning.


\begin{small}
	\textit{Consider the following legal case about Medically Assisted Reproduction where an agent must decide whether it is appropriate this procedure for a specific couple.}
	
	\vspace{-5pt}
	\begin{itemize}
		
		\item[$\mathbb{A}$] \emph{The couple can generate children, then they are not sterile. 
			Furthermore, there exists reasons to think that the couple can grants a reasonably expectancy of life for the child. 
			In conclusion, the couple can not perform the medically assisted reproduction.}
		
		\item[$\mathbb{B}$] \emph{The couple is affected by a serious genetic disease. 
			However, they enjoy psychological well-being. 
			Also, they have access to medically assisted reproduction techniques due to the fact that the woman do not have any physical impediment. 
			Thus, there is a legal solution for the reproduction problems.}
		
		\item[$\mathbb{C}$] \emph{The couple has a stable economic position to provide health and education for the future child.
			Thus, there is a legal solution for the reproduction problems.}		
		
	\end{itemize}
\end{small}
\vspace{-6pt}
{\small \emph{The facts of the case are: the couple is able to conceive and generate children, 
		but they are both carriers of a severe genetic disease, which does not allow children to live for more than a few years. 
		This couple has a good economic position an they enjoys psychological well-being.}}

\smallskip
In order to reach a conclusion for a particular decision, certain features of the knowledge can be taken into account. 
For instance, the \textit{relevance} of the information representing that some pieces of knowledge may be more pertinent than others.
Also the \textit{intuition} behind the knowledge is important. 
Note that these two features are of different nature: one of them (relevance) is a measure usually expressed with real numbers, while the other (intuition) is just a label denoting the implicit or explicit intuition associated with the argument.
Hence, the former is a quantitive feature about the knowledge while the latter is a qualitative one.
Our intuition behind the combination of qualitative and quantitative features associated with arguments is that, in most cases, these features are complementary or with a strong dependency between each other and they are extremely important for legal decision making. 
Thus, by combining these kind of information in the argumentation process, an essential part of legal debates is captured. 

In this work we propose an extension of \emph{Labeled Argumentation Frameworks} (LAF) ~\cite{budan2015modeling,BudanSVS17} in the legal domain that allows the representation of qualitative and quantitative features associated to the arguments involved in a legal dispute. 
In particular, labels are combined and propagated through an argumentation graph according to the manner in which the interactions between arguments are defined: support, conflict, and aggregation. 
Once the propagation process is complete, with the \textit{definitive} argumentation labels, we establish the acceptability status of arguments by using the information on these labels. 

\vspace{-15pt}

\section{Labeled Argumentation Framework}


Labels represent quantitative and qualitative domain-dependant information about the arguments. 
We define an \emph{Algebra of Argumentation Labels} as an abstract algebraic structure that contains the operations related to manipulation of arguments. 
The effect of aggregation, support, and conflict of arguments will be reflected in their labels, evidencing how the arguments have been affected by their interaction. 

\begin{Definition}\label{Def.Algebra}
	An algebra of argumentation labels is a tuple \AForm, where 
	\E\ is a set of labels called the \emph{domain of labels},
	$\le$ is a partial order over \E\ with
	$\top$ and $\bot$ two distinguished elements of \E\,
	$\OR: \E \times \E \rightarrow \E$ is a \emph{support operation},
	$\OA: \E \times \E \rightarrow \E$ is an \emph{aggregation operation}, and
	$\OC: \E \times \E \rightarrow \E$ is a \emph{conflict operation}.
\end{Definition}	



A natural way of representing \textit{quantitative} information is to use a numeric scale. 
We will consider fuzzy valuations ranging between two distinguished elements: \IES\ and \IE, where \IE\ represents the less possible degree, while \IES\ is the maximum degree. 
Regarding \textit{qualitative} features, we are interested here in being as general as possible using \textit{order theory}, 
as a mean for describing statements such as ``\textit{this is less than that}'' or ``\textit{this precedes that}''. 
The appropriate behavior in the argumentation domain is defined by properties associated to the operators.
\smallskip

We extend a formalism called \emph{Labeled Argumentation Framework} (LAF)  that combines the knowledge representation features provided by the \emph{Argument Interchange Format} with  the processing of meta information using the \emph{Algebra of Argumentation Labels}. 
This framework allows a good representation of arguments, by taking into account their internal structure, the argument interactions, and 
special features of the arguments. 
Labels then are propagated and combined through the argumentation process using the algebra operators. 
The final label attached to each argument is obtained, and this information is used to establish the status of acceptance of these arguments.
In LAF, we use the AIF ontology as the underlying knowledge representation model for the internal structure of the arguments and its relations 
\cite{AIF-GR,BudanSVS17}. 

\begin{Definition}\label{Def.LAF}
	A Labeled Argumentation Framework $($LAF$)$ is a tuple of the form $\Phi = \LAFARG$ where: \LLa\ is a logical language for knowledge representation; \R\ is a set of inference rules $R_1, R_2,\ldots, R_n$ defined in terms of \LLa; \K\ is the knowledge base, a set of formulas of \LLa\ describing the knowledge about a specific domain of discourse; \ALL\ is a set of algebras of argumentation labels $\mathsf{A}_1, \mathsf{A}_2,\ldots,$ $\mathsf{A}_n$, one for each feature that will be represented by the labels; and $\F_{\K}$ is a function that assigns to each element of \K\ a $n$-tuple of elements$\,$ in the algebras $\mathsf{A}_i, i=1,\ldots, n$.
\end{Definition}


\begin{Example}\small\label{Example_InstantiateLAF}
	The running example about assisted fertility can be formalized by the following LAF:
	
	\smallskip
	\noindent
	$-$ \LLa\ is a language defined in terms of two disjoint sets: a set of presumptions and a set of defeasible rules, where a presumption is a ground atom $X$ or a negated ground atom $\nega X$, 
	a defeasible rule is an ordered pair, denoted $\argdrule{\mathtt{C}}{\mathtt{P}_1, \ldots,  \mathtt{P}_n}$, where $\mathtt{C}$ is a ground atom (conclusion) and  $\mathtt{P}_1, \ldots ,\mathtt{P}_n$ is a finite non-empty set of ground atoms (premises).
	
	\smallskip
	\noindent				
	$-$ \R = \{\,dMP\}, denoting Defeasible Modus Ponens:
	\medskip
	\hspace*{16pt}dMP:\ {\large $\frac{\mathtt{P}_1, \ldots, \mathtt{P}_n     \     \     \   \mathtt{C} {\scriptstyle\at} \mathtt{P}_1, \ldots,  \mathtt{P}_n} {\mathtt{C}}$} 
	\vspace{-10pt}
	\smallskip
	\noindent
	$- \ALL = \{ \mathsf{A}, \mathsf{B}\}$ the set of algebras of argumentation labels where:
	
	\smallskip
	
	\noindent $\mathsf{A}$ represents the trust degree attached to arguments. The domain of labels $A$ is the real interval $[0,1]$ representing a normalized relevance valuation of the information. 
	
	\smallskip
	
	\noindent	$\alpha \ \OR \ \beta  = \alpha\beta$, where the relevance of a conclusion is based on the conjunction of the relevances corresponding to its premises
	
	\noindent	$\alpha \ \OA \ \beta  = \alpha + \beta - \alpha \beta$, where if there is more than one argument for a conclusion, its relevance valuation is the sum of the valuations of the arguments supporting it, with a penalty term.
	
	\noindent $\alpha \ \OC \ \beta = max(\alpha - \beta,0)$, reflects that the relevance valuation of a conclusion is weakened by the relevance in its contrary.

	\smallskip
	
	\noindent $\mathsf{B}$ is an algebra of argumentation labels representing the intuition attached to arguments. The domain of labels $B$ is the set 
	{\small$\{$ PL (Preserve Life), NG (No Eugenesis), FCH (Preserve the Future of the Child), PCH (Preserve the Couple Health), $\emptyset$ (without intuition)$\}$ }
	representing the intuition behind an argument, where the order of these elements is $PL > NG > FCH > PCH > \bot$. 
	
	\smallskip
	
	\noindent		$\alpha \ \OR \ \beta  = min(\alpha \cup \beta)$, models that the governing intuition for an argument is based on the less influential intuition associated to the arguments that support it (conditional instantiation).
	
	\noindent	$\alpha \ \OA \ \beta  = \alpha \cup \beta$, where if there is more than one argument for a conclusion, its governing intuitions are the composition of the predominant intuition proposed by the arguments supporting it.
	
	\noindent	$\alpha \ \OC \ \beta =	\alpha \setminus \beta$, reflects that only the unquestionable intuition are preserved.
	
	\smallskip
	
	%
	%
	%
	
	\noindent		
	- $\K$ is composed by the set of norms representing the interpretation of the provisions of the legal systems involved in the case, and the intuition behind it.
	We display below the set of formulas of \LLa\ forming \K, where labels (between brackets), denote relevance and intuition behind the formula:\\
	
	\vspace{-15pt}
	
	\begin{footnotesize}
		\begin{center}
			$\left\{
			\begin{array}{l}	
			\tt{r}_1:\argdrule{\tt{Med\_Repr}(X)}{\tt{\nega{Physical\_Imp}}(X)} : 											\{07,PL\}\\[2pt]
			
			
			\tt{r}_2:\argdrule{\tt{\nega{Steril}}(X)}{\tt{Genere\_Child}(X)}:												\{0.7,NE\}\\[2pt]			
			
			\end{array}
			\right\}$
			
			$\left\{
			\begin{array}{l}
			\tt{n}_1:\argdrule{\tt{Sol\_Rep\_Prob}(X)}{\tt{Genetic\_Dis}(X),\tt{Med\_Repr}(X)} :							\{1,PL\}\\[2pt]
			
			
			\tt{n}_2:\argdrule{\tt{\nega{Med\_Repr}}(X)}{\tt{Rsn\_Exp\_Life}(X),\tt{\nega{Steril}(X)}}:						\{0.8,NE\}\\[2pt]
			
			\tt{n}_3:\argdrule{\tt{Sol\_Rep\_Prob}(X)}{\tt{Edu\_Health\_Child(X)}}:									\{0.5,FCH\}\\[2pt]
			\end{array}
			\right\}$

			$\left\{
			\begin{array}{ll}
			\facto{\tt{\nega{Physical\_Imp}}(CP)} : \{0.8,PL\}			& \facto{\tt{Genere\_Child}(CP)} : \{0.8,NE\}	 \\[2pt]	
			\facto{\tt{Rsn\_Exp\_Life}(CP)}: \{0.5,PL\}		   		& \facto{\tt{Genetic\_Dis}(CP)} :\{1,PL\} \\[2pt]	
			\facto{\tt{Edu\_Health\_Child(CP)}} : \{0.6,FCH\}	& \\[2pt]
			\end{array}
			\right\}$
			
		\end{center}
	\end{footnotesize}
	
\end{Example}

\vspace{-10pt}

We use \textit{argumentation graphs} to represent the argumentative analysis derived from an \textit{LAF}. We assume that nodes are named with different sentences of \LLa, so we will use the naming sentence to refer to the I-node in the graph.
\begin{Definition}~\label{Def.ArgumentationGraph}
	Given an LAF $\Phi$, its associated  argumentation graph is the digraph $\Graph{\Phi} = (N, E)$, where $N \neq \emptyset$ is the set of nodes and $E$ is the set of the edges, constructed as follows: 	(i) each element $X \in \K$ or derived from $\K$ through \R, is represented by an I-node $X \in N$. (ii) for each application of an inference rule defined in $\Phi$, there exists an RA-node $R \in N$ such that: (a) the inputs are all I-nodes $P_1,\ldots, P_m \in N$ representing the premises necessary for the application of the rule $R$; and (b) the output is an I-node $Q \in N$ representing the conclusion. (iii) if $X$ and $\negaBar{X}$ are in $N$, then there exists a CA-node with edges to and from both I-nodes $X$ and $\negaBar{X}$.	(iv) for all $X \in N$ there does not exist a path from $X$ to $X$ in $G_{\Phi}$ that does not pass through a CA-node. 
	
	%
	%

\end{Definition}

Once the argumentation graph is obtained, we proceed to attach a label to each I-node, representing the features referring to extra information that we want to represent. 

\begin{Definition}~\label{Def.Labelgraphcycle}
	Let $\Phi$ be an LAF, and $\Graph{\Phi}$ be its corresponding argumentation graph.
	Let $\mathsf{A}_i$ be one of the algebras in $\mathcal{A}$.
	A labeled argumentation graph is an assignment of two valuations from each of the algebras to all I-nodes of the graph, denoted with \Labelplus{i}{X} and \Labelminus{i}{X}, 
	where \Labelplus{i}{X} accounts for the aggregation of the reasons supporting the claim $X$, 
	while \Labelminus{i}{X} displays the state of the claim after taking conflict into account, such that \Labelplus{i}{X}, \Labelminus{i}{X} $\in \mathit{A}_i$. 
	If $X$ is an I-node, its valuations are determined as follows: 	(i) If $X$ has no inputs, then its accrued valuation is given by function \F; thus, $\Labelplus{i}{X} = \F_i(X)$. (ii) If $X$ has an input from a CA-node representing conflict with $\negaBar{X}$, then: $\Labelminus{i}{X} =\Labelplus{i}{X} \Con\ \Labelplus{i}{\overline{X}}$. If there is no input from a CA-node then: $\Labelminus{i}{X} = \Labelplus{i}{X}$. (iii) If $X$ is an element of $\K$ with inputs from RA-nodes $R_1, \ldots, R_k$, where each $R_s$ has premises $X_{1}^{R_s}, \ldots, X_{n_s}^{R_s}$, then: $\Labelplus{i}{X} = \F_i(X) \Agr [\Agr^k_{s=1}(\Sup^{n_s}_{t=1} \Labelminus{i}{X_t^{R_s}})]$. If $X$ is not an element of $\K$ and has inputs from RA-nodes $R_1, \ldots, R_k$, where each $R_s$ has premises $X_{1}^{R_s}, \ldots, X_{n_s}^{R_s}$, then:	$\Labelplus{i}{X} = \Agr^k_{s=1}(\Sup^{n_s}_{t=1} \Labelminus{i}{X_t^{R_s}})$.
	
	%
	%
	%
	%

\end{Definition}

This procedure determines the system of equations representing the constraints that all features must fulfill.
Once the I-nodes of the argumentative graph are labeled, we can determine the acceptability status associated with an argument.

\begin{Definition}\label{AcceptabilityDegreeStatus}
	Let  $\Phi$ be an LAF. 
	For each of the algebras $\mathsf{A}_i$ in $\mathcal{A}$, representing a feature to be associated with each I-node $X$.
	Then, $X$ has assigned one of four possible acceptability statuses with respect to $\mathsf{A}_i$: $\mathtt{Assured}$ if and only if $\Labelminus{i}{X} = \neutros_i$ or $\neutros_i \in \Labelminus{i}{X}$; $\mathtt{Unchallenged}$ if and only if $\Labelplus{i}{X} = \Labelminus{i}{X} \neq \bot_i$; $\mathtt{Weakened}$ if and only if $\neutro_i < \Labelminus{i}{X} < \Labelplus{i}{X}$; and $\mathtt{Rejected}$ if and only if $\Labelminus{i}{X} = \neutro_i$.
	
	
\end{Definition}

Finally, for each claim, we form a vector with the acceptability of that claim with respect to each of the attributes, and take the least degree of those that appear in the vector as the acceptability degree for the claim as a whole. 

\begin{Example}\small\label{examplelabeledGraph}
	We show in Figure~\ref{Fig.argumentationgraphlabeles} the solution to EQS, which represent the model of the instantiated LAF. For reasons of readability and space, we do not include the equations that determine the features of the graph.
	
	%
	%
	
	\begin{figure}[h]
		\centering
		\includegraphics[width=1\textwidth]{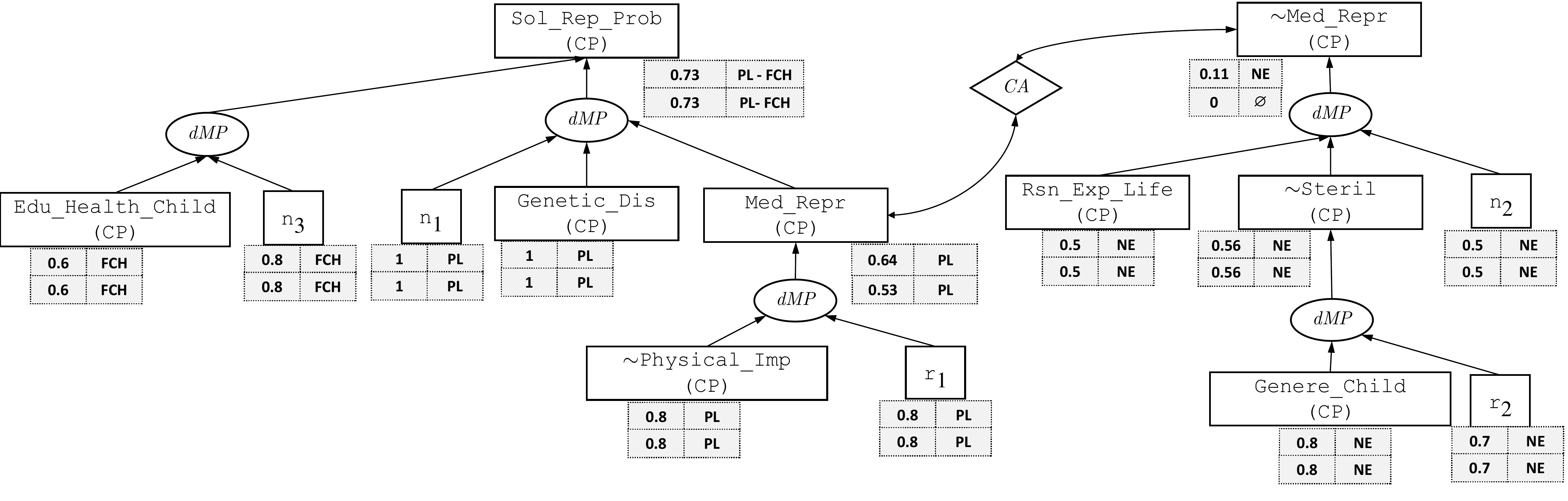}
		\caption{Labeled for an Argumentation graph}\label{Fig.argumentationgraphlabeles}
	\end{figure}
	
	\vspace{-10pt}
	
	\noindent We calculate the acceptability status for each claim, considering the valid labeling for the argumentation graph that describes the discussion about medically assisted reproduction. Thus, we have that $S^A_g = \{\tt{Genetic\_Dis}(CP), \tt{n}_1\}$, $S^U_g = \{\nega{\tt{Physical\_Imp}(CP)}, \tt{Edu\_Health\_Child}(CP),$ $\tt{Genere\_Child}(CP), \tt{Rsn\_Exp\_Life}(CP), \nega{\tt{Steril}(CP)}, \tt{n}_2, \tt{n}_3, \tt{r}_1, \tt{r}_2\}$, $S^W_g = \{\tt{Med\_Repr}(CP)\}$, and $S^R_g = \{\nega{\tt{Med\_Repr}(CP)}\}$.
	Finally, the medically assisted reproduction is legally accepted with a relevance of 0.72/), and a intuition of preserve the life and the future of the child.
\end{Example}
\vspace{-25pt}

\section{Conclusions}\label{Conclusion_work}

In this work we present an extension of  \emph{Labeled Argumentation Framework} that combines the KR capabilities provided by the \emph{Argument Interchange Format} (\emph{AIF}), together with the management of labels through properly defined algebras. 
We have associated operations in an algebra of argumentation labels to three different types of argument interactions,  allowing to propagate qualitative and quantitative information on the argumentation graph. 
We defined a process to determine the status of acceptance of arguments, by integrating the information shown in labels. 
The conflict operation defined in the algebra allows the weakening of arguments, which contributes to a better representation of application domains.
In our running example, labels are used to characterize relevance and the intuition associated with the arguments supporting decision making. 
These qualitative and quantitative features are the key to provide a properly, founded decision by providing awareness of different aspects of the knowledge exposed.

\noindent
\vspace{-15pt}
\bibliographystyle{plain}

\bibliography{bib}

\end{document}